\documentclass[letterpaper, 10 pt, conference]{ieeeconf}  

\IEEEoverridecommandlockouts                              

\usepackage{graphicx}
\usepackage{subfigure} 
\usepackage{amsmath,amsfonts}
\usepackage{multirow} 
\usepackage{makecell} 
\usepackage[table,svgnames]{xcolor} 
\usepackage{colortbl} 
\usepackage{diagbox} 
\usepackage{arydshln} 
\usepackage{tabularx}
\usepackage{booktabs} 

\usepackage{hyperref}
\usepackage{cleveref}

\colorlet{colorFst}{LightGreen}       
\colorlet{colorSnd}{cyan} 
\colorlet{colorTrd}{LightGrey!65}      
\colorlet{colorLow}{darkgray!30}    

\colorlet{cellcolorFst}{LightGreen!35}       
\colorlet{cellcolorSnd}{LightBlue!75} 
\colorlet{cellcolorTrd}{LightGrey!65}      

\newcommand{\fs}[1]{{\textbf{#1}}}   
\newcommand{\nd}[1]{{#1}}     

\newcommand\ours{NeuraLoc}
\newcommand\myvspace{\vspace{0mm}}

\title{\LARGE \bf \ours: Visual Localization in Neural Implicit Map with Dual Complementary Features}

\author{Hongjia Zhai$^{1}$, Boming Zhao$^{1}$, Hai Li$^{2}$, Xiaokun Pan$^{1}$, Yijia He$^{2}$, Zhaopeng Cui$^{1}$, \\ Hujun Bao$^{1}$, Guofeng Zhang$^{1*}$
\\ $^{1}$State Key Lab of CAD\&CG, Zhejiang University \quad $^{2}$RayNeo
\thanks{* Corresponding author: Guofeng Zhang (zhangguofeng@zju.edu.cn)}
\thanks{This work was partially supported by the NSF of China (No.6242500063).}
}

\begin{document}

\maketitle
\thispagestyle{empty}
\pagestyle{empty}

\begin{abstract}
Recently, neural radiance fields (NeRF) have gained significant attention in the field of visual localization. However, existing NeRF-based approaches either lack geometric constraints or require extensive storage for feature matching, limiting their practical applications. To address these challenges, we propose an efficient and novel visual localization approach based on the neural implicit map with complementary features. Specifically, to enforce geometric constraints and reduce storage requirements, we implicitly learn a 3D keypoint descriptor field, avoiding the need to explicitly store point-wise features. To further address the semantic ambiguity of descriptors, we introduce additional semantic contextual feature fields, which enhance the quality and reliability of 2D-3D correspondences. Besides, we propose descriptor similarity distribution alignment to minimize the domain gap between 2D and 3D feature spaces during matching. Finally, we construct the matching graph using both complementary descriptors and contextual features to establish accurate 2D-3D correspondences for 6-DoF pose estimation. Compared with the recent NeRF-based approaches, our method achieves a 3$\times$ faster training speed and a 45$\times$ reduction in model storage. Extensive experiments on two widely used datasets demonstrate that our approach outperforms or is highly competitive with other state-of-the-art NeRF-based visual localization methods. Project page: \href{https://zju3dv.github.io/neuraloc}{https://zju3dv.github.io/neuraloc}
\end{abstract}

\section{INTRODUCTION}
Visual localization plays a crucial role in many robotics applications~\cite{ming2024benchmarking,zhu:2021:niceslam,nerf-loc}, aiming to estimate the 6-Degree-of-Freedom (6-DoF) camera pose of a query image with respect to a pre-built 3D map.

With the advancement of deep learning, convolutional neural networks (CNNs) have shown great potential in extracting high-level contextual features, significantly promoting the development of visual localization. Current visual localization approaches can be broadly categorized into feature-based and regression-based methods.
Feature-based methods~\cite{hloc,bgnet} typically represent the scene using point clouds. 
These methods rely on keypoint detection and matching techniques~\cite{superpoint,superglue,r2d2,sift} to reconstruct a 3D point map. 
Once 2D-3D correspondences between the query image and the point-cloud map are established, the 6-DoF pose can be estimated via the Perspective-n-Point (PnP) algorithm~\cite{epnp}. 
However, their performance is limited by the repeatability and discriminative power of the extracted or learned keypoint descriptors. Furthermore, feature-based methods require substantial storage to retain map information, such as 3D points, descriptors, and covisibility relationships, which can be storage-intensive.
On the other hand, regression-based approaches~\cite{mapnet,posenet,scrnet} employ CNNs to directly regress the pose from features extracted from a single image or to estimate the 3D coordinates of pixels in the camera's view. While these methods avoid the need for storing point-wise descriptors and demand less storage space, their generalization performance in large, complex scenarios is relatively poor. Additionally, regression-based methods typically require large amounts of 3D data to optimize the regression network. As a result, while these methods are more storage-efficient, they are generally inferior to feature-based methods in terms of localization accuracy and generalization performance.

Recently, NeRF~\cite{mildenhall:2020:nerf} has emerged as a new paradigm for implicit scene representation.
The neural implicit representation uses multi-layer-perceptions (MLPs) and additional parametric encoding~\cite{mueller2022instantngp,tetra-nerf,Vox-Surf,eslam,imtooth} to model the scene property and achieve impressive results in novel view synthesis and surface reconstruction.
Benefiting from differentiable volume rendering~\cite{volume_rendering}, NeRF-based approaches enable end-to-end parameter optimization without the need of 3D supervision.
Several recent works~\cite{nerf-loc,inerf,pnerfloc,uncertainty-loc,nerf-loc-transformer} have explored using neural implicit representations to support 6-DoF pose estimation and refinement. For example, iNeRF~\cite{inerf} was the first method to use a pre-trained NeRF model for estimating the pose of an unknown query image. However, it is limited to object-level localization and requires an initial pose estimate.
In visual localization, LENS~\cite{lens} and NeRF-SCR~\cite{uncertainty-loc}  are regression-based approaches that use NeRF to synthesize virtual camera views to aid scene coordinate regression networks. 
However, similar to other regression-based methods, these approaches suffer from unsatisfactory localization results due to the lack of geometric constraints.
To introduce geometric constraints, PNeRFLoc~\cite{pnerfloc} uses point-based neural representation~\cite{xu2022pointnerf} to represent the scene map and performs 2D-3D feature matching for pose estimation.
This was the first work to combine feature-based methods with NeRF-based representations for visual localization. While PNeRFLoc achieves satisfactory results across various scenes, it requires explicit storage of features for each point, leading to significant storage requirements for the reconstructed scene model.

To address the issues mentioned above, we propose an efficient and novel visual localization approach, \ours, based on a learned neural implicit map. Specifically, to reduce the size of the scene model, we avoid explicitly storing point cloud descriptors. Instead, we implicitly learn a descriptor field from an MLP decoder, distilled from a 2D keypoint detection model~\cite{superpoint}. Additionally, to address the semantic ambiguity of the learned descriptors and filter outliers, we distill a semantic contextual feature field, which helps construct the accurate matching graph. Furthermore, to reduce the domain gap between 2D and 3D descriptor feature spaces, we introduce a similarity alignment loss to align 2D-3D and 2D-2D similarity distributions. Finally, we construct a matching graph using the dual complementary features to establish 2D-3D correspondences for 6-DoF pose estimation.

In summary, the major contributions of our proposed approach are summarized as follows:
\begin{itemize}
    \item{We propose an efficient visual localization approach based on a reconstructed neural implicit map, achieving accurate localization with fewer scene model parameters.}
    \item To maintain localization performance with a compact scene model, we introduce dual complementary descriptors and semantic contextual features distilled from 2D foundation models for accurate 6-DoF pose estimation.
    \item{We reduce the domain gap between 2D and 3D feature spaces during matching by applying alignment loss on descriptor similarity distributions.}
\end{itemize}

\section{RELATED WORKS}
\subsection{Visual Localization}
Visual localization approaches can be broadly divided into regression-based~\cite{mapnet,posenet,dsac}, and feature-based methods~\cite{hloc,pnerfloc,bd_loc}.
Regression-based approaches often rely on CNNs to directly regress either the camera pose~\cite{mapnet} or the scene coordinates~\cite{src}. One of the key limitations of these methods is their heavy dependence on large amounts of 3D ground truth data for supervision. Additionally, they generally lack interpretability, which limits their performance in terms of both localization accuracy and generalization.
On the other hand, feature-based methods~\cite{hloc,bgnet}, which are the most commonly used for visual localization, utilize sparse or dense feature matching to estimate camera poses by establishing 2D-3D correspondences.
Thanks to advanced keypoint detection~\cite{superpoint,r2d2} and matching techniques~\cite{superglue,loftr}, feature-based approaches can reliably compute 2D-3D correspondences between query images and a prebuilt 3D map. For pose estimation, these methods typically apply Perspective-n-Point~\cite{epnp} and RANSAC~\cite{ransac} algorithms to register query images.
Regression-based methods tend to have fewer parameters but suffer from lower generalization ability and localization accuracy compared to feature-based methods.

\subsection{Neural Radiance Fileds}
Recent advances in implicit representations have shown great potential in various 3D computer vision tasks, such as surface reconstruction~\cite{wang:2021:neus,oechsle:2021:unisurf,azinovic:2022:neuralrgbd}, and neural implicit SLAM~\cite{zhu:2021:niceslam,sucar:2021:imap,vox-fusion,co-slam,nis_slam,vox-fusion++}.
For example, NeuS~\cite{wang:2021:neus} employs a signed distance function (SDF) to represent object geometry and optimizes it using volumetric rendering techniques~\cite{volume_rendering} with an unbiased weighting function.
Similarly, iMAP~\cite{sucar:2021:imap} is the first neural SLAM approach to use MLPs to represent the scene, performing bundle adjustments to optimize the tracked RGB-D camera trajectory.
However, MLPs have limited representation capacity when applied to large-scale reconstructions, and additional parametric encoding features are proposed, such as voxel~\cite{Vox-Surf,liu:2020:nsvf}, point~\cite{xu2022pointnerf}, and tri-plane~\cite{eslam}.
To extend implicit representations to outdoor scenes, Tao recently proposed SILVR~\cite{tao2024silvr}, the first neural Lidar SLAM system, capable of handling large-scale, outdoor environments.

\subsection{NeRF-based Visual Localization}
Recently, several approaches~\cite{nerf-loc,pnerfloc,uncertainty-loc,nerf-loc-transformer,lens,loc-nerf,splatloc} have explored the use of differentiable representation for pose estimation and visual localization, benefiting from their high-quality novel view synthesis capabilities.
For instance, iNeRF~\cite{inerf} was the first to estimate the camera pose of a query image by leveraging the differentiability of a pre-trained NeRF. However, iNeRF requires a coarse initial camera pose and is primarily limited to object-level pose estimation. LENS~\cite{lens} employs NeRF-W~\cite{martin2021nerfw} to reconstruct scenes and synthesize virtual camera views, aiding the training of a scene coordinate network. 
To address the need for geometric constraints, methods such as~\cite{nerf-loc,pnerfloc} utilize robust keypoint detection~\cite{r2d2} and matching networks~\cite{loftr} to improve pose estimation. Zhao introduces PNeRFLoc~\cite{pnerfloc}, which adapts point-based neural representations to facilitate both feature matching and rendering, further enhancing the localization process.

\section{METHOD}
\begin{figure*}[ht!]
\centering
 \includegraphics[width=\linewidth]{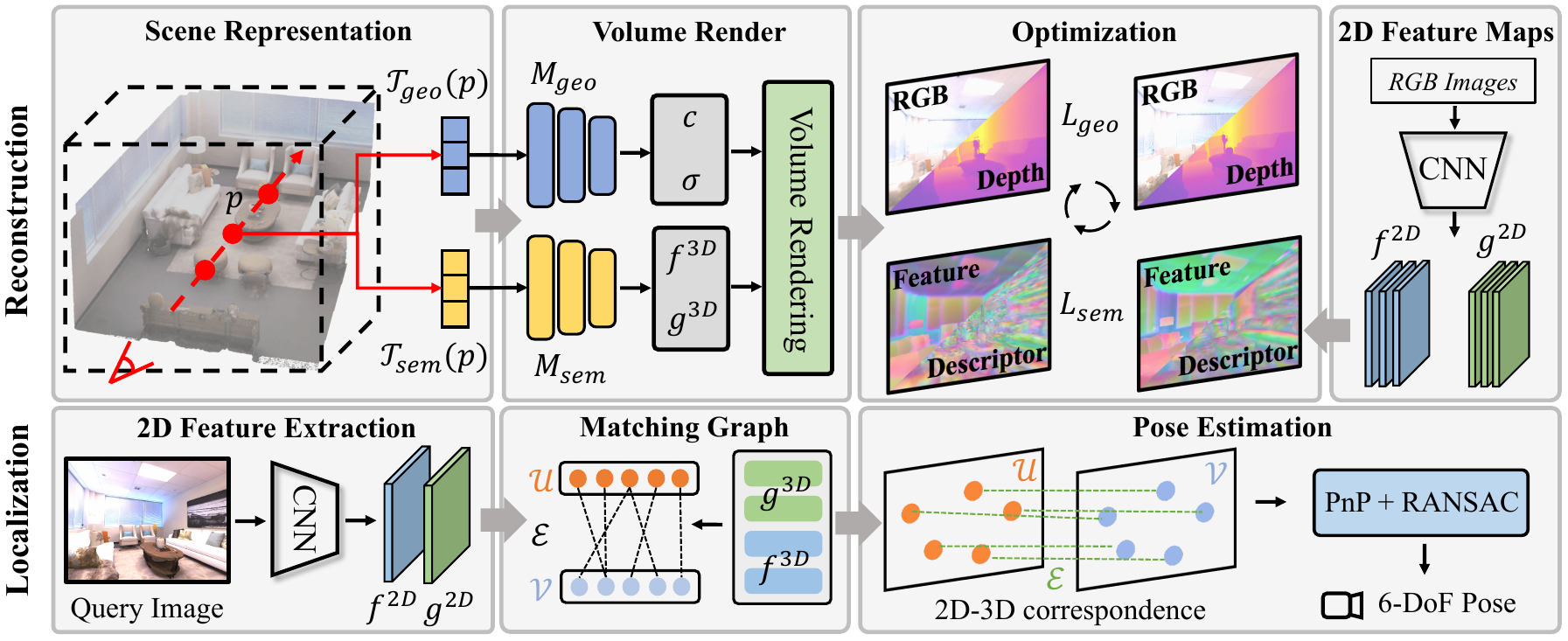} 
\caption{The whole pipeline of our system. (1) Reconstruction: We employ different parametric encodings ($\mathcal{T}_{geo}$ and $\mathcal{T}_{sem}$) for geometry and semantic branches. Scene properties, including color $c$, SDF $\sigma$, semantic contextual feature $f^{3D}$, and keypoint descriptor $g^{3D}$ are produced by separated shadow decoders ($\mathcal{M}_{geo}$ and $\mathcal{M}_{sem}$). We use pre-trained CNN models (SuperPoint~\cite{superpoint} and SAM~\cite{sam}) to generate 2D feature maps for the optimization of the semantic branch. (2) Localization: We extract 2D descriptors and semantic contextual features for the query image to build the matching graph between 3D points. Then, we estimate the 6-DoF pose based on the 2D-3D correspondence.}
\label{fig:pipeline}
\end{figure*}

Given posed RGB-D images $\{I_i\in\mathbb{R}^3, D_i\in\mathbb{R},\}$, we first extract 2D feature maps with powerful vision foundation models~\cite{superpoint,sam} for RGB images.
Then, we reconstruct the consistent appearance/geometry property and dual complementary descriptor/contextual feature fields for the entire scene (Sec.~\ref{subsec:reconstruction}).
With the reconstructed neural implicit semantic map, we can estimate the 6-DoF pose of the query image based on the matching graph (Sec.~\ref{subsec:localization}).
The whole reconstruction and localization pipeline is shown in Fig.~\ref{fig:pipeline}.

\subsection{Reconstruction Process}
\label{subsec:reconstruction}
In this part, we introduce the geometry and semantic reconstruction process of our neural implicit map.

\myvspace\noindent\textbf{Scene Representation and Volume Rendering.}
As noted in~\cite{sucar:2021:imap}, a single MLP struggles with representing large scenes.
Therefore, to improve the MLP's representation capacity, we employ the multi-resolution hash encoding~\cite{mueller2022instantngp} to capture high-frequency information (\textit{e.g.}, color, contextual feature).
As illustrated in the top part of Fig.~\ref{fig:pipeline}, we utilize separate hash parametric encodings for different branches, $\mathcal{T}_{geo}=\{\mathcal{T}_{geo}^{l}\}_{l=1}^{L}$ for geometry branch and $\mathcal{T}_{sem}=\{\mathcal{T}_{sem}^{l}\}_{l=1}^{L}$ for semantic branch, $L$ is the level of different resolutions.
For each sampled point, $p\in\mathbb{R}^{3}$, along the camera ray, its encoding feature is obtained via trilinear interpolation from the multi-resolution feature volume.
To decode the scene properties, we adopt two separate shallow MLPs:
\begin{equation}
    (c,\sigma),(f^{3D},g^{3D}) = \mathcal{M}_{geo}(\mathcal{T}_{geo}(p)), \mathcal{M}_{sem}(\mathcal{T}_{sem}(p)),
\end{equation}
Here, $\mathcal{M}_{geo}$ and $\mathcal{M}_{sem}$ are the MLP decoders for each branch. where $c$, $\sigma$ are the color, SDF, and $f^{3D}$, $g^{3D}$ are the learned 3D semantic contextual feature and descriptors.

Instead of using occupancy or density-based rendering~\cite{mildenhall:2020:nerf,martin2021nerfw}, we adopt SDF-based volume rendering~\cite{wang:2021:neus,azinovic:2022:neuralrgbd,vox-fusion} for superior scene geometry reconstruction.
Following~\cite{azinovic:2022:neuralrgbd}, we adopt a simple bell-shaped rendering weight function that can transform the signed distance function into rendering weight:
\begin{equation}
   \quad w_i = \text{Sigmoid}(\frac{\sigma_i}{tr})\cdot \text{Sigmoid}(-\frac{\sigma_i}{tr}),
\end{equation}
where $tr$ is the truncation distance for supervising $\sigma$.

The scene and semantic properties of each ray are then obtained through volume rendering~\cite{volume_rendering} with the following equation:
\begin{equation}
    c= \sum_{i=1}^{M} \widetilde{w}_{i} c_i, \quad d= \sum_{i=1}^{M} \widetilde{w}_{i} d_i,
    \label{eq:render_geo}
\end{equation}
\begin{equation}
    f^{3D} = \sum_{i=1}^{M} \widetilde{w}_{i} f^{3D}_i, \quad g^{3D} = \sum_{i=1}^{M} \widetilde{w}_{i} g^{3D}_i,
    \label{eq:render_sem}
\end{equation}
where $M$ is the number of sampled pixels along each ray, and $\widetilde{w}_{i}$ is the normalized weight, calculated as $\widetilde{w}_i=w_i/\sum_j^M w_j$.

\myvspace\noindent\textbf{Scene Reconstruction.}
To reconstruct the appearance and geometry of the scene, we apply the loss functions like~\cite{azinovic:2022:neuralrgbd}: RGB loss ($L_{c}$), Depth loss ($L_{d}$), SDF loss ($\mathcal{L}_{sdf}$), and free space loss ($\mathcal{L}_{fs}$). Those losses are defined as follows:
\begin{equation}
    \mathcal{L}_{c} = \sum_{r\in R}(c(r) - I(r))^2, \mathcal{L}_{d} = \sum_{r\in R}(d(r) - D(r))^2
\end{equation}
where $\mathcal{L}_{c}$ and $\mathcal{L}_{d}$ are performed for each sampled ray $r$.
For more accurate geometry learning, $\mathcal{L}_{fs}$ and $\mathcal{L}_{sdf}$ are used  to optimze the signed distance value, $\sigma$:
\begin{equation}
    \mathcal{L}_{fs} = \sum_{r\in R} \frac{1}{|P_r^{fs}|}\sum_{p \in P_r^{fs}} (\sigma - 1)^2,
\end{equation}
\begin{equation}
    \mathcal{L}_{sdf} = \sum_{r\in R} \frac{1}{|P_r^{tr}|}\sum_{p \in P_r^{tr}} (d(p) + \sigma \cdot tr - D(r))^2,
\end{equation}
where $P_r^{fs}$ and $P_r^{tr}$ are the point set which outside the truncation region, $|d(p) - D(r)| > tr$, and inside the truncation region, $|d(p) - D(r)| < tr$.

The final reconstruction loss for the geometry branch is expressed as:
\begin{equation}
    \mathcal{L}_{geo} = \lambda_{c} \cdot \mathcal{L}_{c} + \lambda_{d} \cdot \mathcal{L}_{d} + \lambda_{sdf} \cdot \mathcal{L}_{sdf} + \lambda_{fs} \cdot \mathcal{L}_{fs},
\end{equation}
where $\lambda_{i}$ are the weights for different components, which are set to 0.5, 1.0, 5000, and 10 respectively.

\myvspace\noindent\textbf{Complementary Feature and Descriptor Distillation.}
To reduce the storage requirements of keypoint descriptors, we learn a discriminative descriptor field in the semantic branch through distillation from the pre-trained 2D keypoint detection model, SuperPoint~\cite{superpoint}.
The learned descriptor can be used to estimate 2D-3D correspondence.
However, due to the semantic ambiguity of similar structures, a single descriptor is not effective enough for localization, which may lead to incorrect correspondence.
To address this, we distill an additional complementary semantic feature field in the semantic branch, focusing on contextual information around pixels. 
We use the segmentation foundation model, SAM~\cite{sam} as the supervision, leveraging its robust capability to identify semantically similar pixels.
To learn 3D consistent context semantic features and keypoint descriptor fields, we first extract the 2D feature map of database images through~\cite{superpoint,sam}.
For each sampled pixel during volume rendering, we obtain its feature in the 2D feature map, $f^{2D}_i$, $g^{2D}_i$.
We align the 3D feature rendered from neural fields to the feature maps with the following equation:
\begin{equation}
    \mathcal{L}_{dis} = \sum_{i} [1 - cos(f^{2D}_i,f^{3D}_i)] + [1 - cos(g^{2D}_i,g^{3D}_i)],
\end{equation}
where $cos(\cdot)$ is the cosine similarity function between two feature vectors, $f^{3D}_i$ and $g^{3D}_i$ are the rendered 3D complementary feature via Eq.~(\ref{eq:render_sem}).

\begin{figure}[ht!]
\centering
\includegraphics[width=\linewidth]{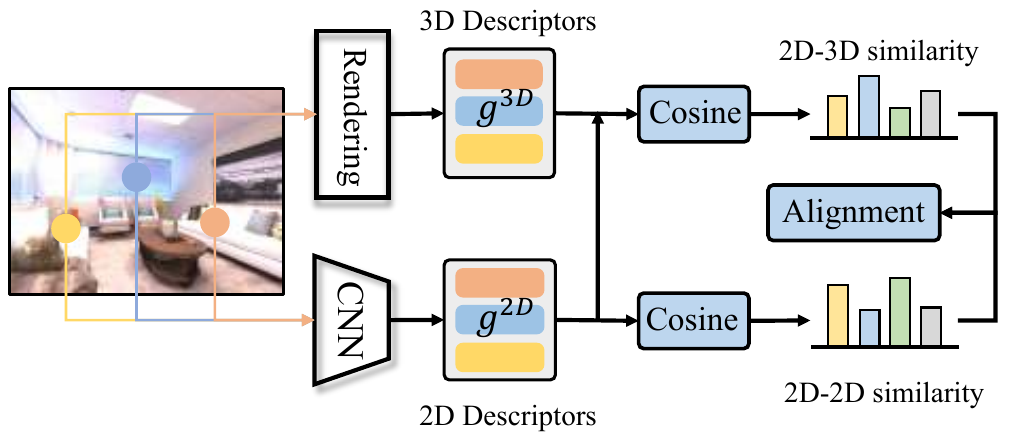}
\caption{Descriptor similarity alignment. To reduce the domain gap, we perform the similarity distribution alignment between the 2D-2D and 2D-3D similarity distribution for better optimization.}
\label{fig:sim_align}
\end{figure}
\myvspace\noindent\textbf{Descriptor Similarity Distribution Alignment.}
We utilize the similarity between 2D and 3D descriptors for feature matching and correspondence estimation. However, due to the inherent domain gap between 2D and 3D feature spaces, this can result in incorrect matches.
As shown in Fig.~\ref{fig:sim_align}, we address this issue by imposing alignment constraints on the 2D-3D similarity distributions.

During each iteration, we randomly sample $M$ keypoints from a selected database frame. First, we obtain the L2-normalized 2D and 3D descriptors, $g^{2D}$ and $g^{3D}$.
For the $i$-th keypoint, we compute its 2D-2D similarity distribution relative to other keypoints in the same image, denoted as $p_{i}^{2D}=\{cos(g^{2D}_i,g^{2D}_j\}_{j=1}^{M}$
Similarly, we calculate the 2D-3D similarity distribution for its 3D descriptor, $p_{i}^{3D}=\{cos(g^{3D}_i,g^{2D}_j\}_{j=1}^{M}$.
Since the relationship between different descriptors is crucial during the matching process, we optimize the 3D descriptor field by aligning the learned 2D-3D descriptor similarity distribution with the original 2D-2D similarity distribution. 
To reduce the gap, we align the similarity distribution with the following equation:
\begin{equation}
    \mathcal{L}_{kl} = \sum_{i} KL(\texttt{Softmax}(p_{i}^{2D}) || \texttt{Softmax}(p_{i}^{3D})),
\end{equation}
where $\texttt{KL}(\cdot)$ is the Kullback-Leibler divergence between two descriptor similarity distributions. Following the approach~\cite{kd}, we apply temperature scaling to focus the optimization on positive matches.

Finally, the reconstruction loss for the semantic branch is formulated as:
\begin{equation}
    \mathcal{L}_{sem} = \lambda_{dis} \cdot \mathcal{L}_{dis} + \lambda_{kl} \cdot \mathcal{L}_{kl},
\end{equation}
where $\lambda_{i}$ are the weights for different components, which are set to 1.0 and 0.01, respectively.

\subsection{Localization Process}
\label{subsec:localization}
In this part, we show the process of localization for query images. The whole process is shown at the bottom of Fig.~\ref{fig:pipeline}.

\myvspace\noindent\textbf{Problem Formulation.}
To obtain the 6-DoF pose of the query image, we need to establish 2D-3D correspondence between the query image and the 3D neural implicit map.
So, this localization problem can be modeled as constructing and solving a bipartite graph matching problem.
The bipartite graph is denoted as $\mathcal{G}=\{\mathcal{U}, \mathcal{V}, \mathcal{E}\}$, where $\mathcal{U}=\{u_1, u_2, \cdots, u_m\} \in \mathbb{R}^{m\times2}$ is the 2D keypoints in the query image, and $\mathcal{V}=\{v_1,v_2, \cdots,v_n\} \in \mathbb{R}^{n\times3}$ is the 3D points in the implicit map.
The edge set $\mathcal{E}=\{e_{i,j}\} \in \mathbb{R}^{m\times n}$, where $e_{i,j}=e(u_i,v_j)$ represent the probability that $u_i$ and $v_j$ are correspond.
The goal of the localization process is to solve the bipartite graph matching problem and obtain the assignment matrix, $\mathcal{S}$, which represents the true correspondences.

\myvspace\noindent\textbf{Matching Graph Construction.}
To construct the graph, we first extract the 2D keypoints $\mathcal{U}$, 2D semantic contextual features $\{f^{2D}_i\}$, and descriptors $\{g^{2D}_i\}$ for each query image using~\cite{superpoint,sam}.
For the 3D points $\mathcal{V}$, we consider those within the frustum of the retrieved reference image~\cite{netvlad} to build the matching graph $\mathcal{G}$.
Their 3D semantic contextual feature $\{f^{3D}_i\}$, and descriptor $\{g^{3D}_i\}$ via volume rendering (Eq. (\ref{eq:render_sem})).
Since there is a domain gap between 2D and 3D feature spaces, directly matching keypoints with descriptors can often lead to incorrect results. To address this, we incorporate the semantic contextual features and compute the 2D-3D matching scores using the following equation:
\begin{equation}
    e_{ij} = \cos(g^{2D}_i, g^{3D}_i) + \alpha \cdot\cos(f^{2D}_i, f^{3D}_j).
\end{equation}
Additionally, to reduce the number of candidate matches in the graph, we filter out matches whose semantic contextual feature similarity falls below a specified threshold.

\myvspace\noindent\textbf{Pose Estimation.}
Inspired by~\cite{bgnet,liu2023matcher}, after constructing the graph $\mathcal{G}=\{\mathcal{U},\mathcal{V},\mathcal{E}\}$, we apply the Hungarian algorithm~\cite{kuhn1955hungarian} to solve the bipartite matching problem and obtain the final matched correspondences.
\begin{equation}
    \mathcal{S} = \texttt{Hungarian}(\mathcal{G}=\{\mathcal{U},\mathcal{V},\mathcal{E}\}),
\end{equation}
where $\mathcal{S}=\{s_1,s_2,\cdots,s_k\}$ is the assignment matrix, with $s_i\in\{0,1\}$ represents whether the $i$-th edge is contained in the maximum-weight matching.
Using the estimated 2D-3D correspondences, we apply the RANSAC and PnP algorithms~\cite{ransac,epnp} to estimate the 6-DoF pose of the query image.


\begin{table*}[h]
\centering
\caption{Visual localization results on Replica dataset. We report median translation and rotation errors (cm, degree).}
\setlength{\tabcolsep}{6.2pt}
\begin{tabularx}{0.9\linewidth}{lcccccccc}
\toprule
Method & Room 0 & Room 1 & Room 2 & Office 0 & Office 1 & Office 2 & Office 3 & Office 4  \\
\midrule
SCRNet~\cite{scrnet} & 2.05 / 0.33 & 1.84 / 0.34 & 1.31 / 0.26 & 1.69 / 0.34 & 2.10 / 0.52 & 2.21 / 0.41 & 2.13 / 0.37 & 2.25 / 0.43 \\
SCRNet-ID~\cite{scrnet-id} & 2.33 / 0.28 & 1.83 / 0.35 & 1.78 / 0.29 & 1.79 / 0.37 & 1.65 / 0.42 & 2.07 / 0.37 & 1.79 / 0.28 & 2.42 / 0.35 \\
SRC~\cite{src} & 2.78 / 0.54 & 1.92 / 0.35 & 2.97 / 0.63 & 1.45 / 0.30 & 2.07 / 0.53 & 2.53 / 0.51 & 3.44 / 0.63 & 4.84 / 0.90\\
NeRF-SCR~\cite{uncertainty-loc} & 1.53 / 0.24 & 1.96 / 0.31 & 1.34 / 0.22 & 1.61 / 0.35 & 1.54 / 0.44 &  1.69 / 0.33 & 2.40 / 0.38 & 1.69 / 0.32 \\
PNeRFLoc~\cite{pnerfloc} & \nd{1.00} / \nd{0.21} & \nd{1.32} / \nd{0.28} & \nd{1.43} / \nd{0.29} & \nd{0.72} / \nd{0.15} & \nd{1.08} / \nd{0.28} & 1.71 / 0.37 & \nd{2.39} / \nd{0.30} & \nd{1.63} / \nd{0.32} \\
Ours & \fs{0.51} / \fs{0.08} & \fs{1.06} / \fs{0.20} & \fs{1.11} / \fs{0.22} & \fs{0.39} / \fs{0.08} & \fs{0.82} / \fs{0.21} & \fs{1.18} / \fs{0.22} & \fs{1.32} / \fs{0.21} & \fs{1.05} / \fs{0.17} \\
\bottomrule
\end{tabularx}
\label{tab:replica}
\end{table*}

\begin{table*}[h]
\centering
\setlength{\tabcolsep}{4.5pt}
\caption{Visual localization results on 12-Scenes dataset. We report median translation and rotation errors (cm, degree).}
\begin{tabularx}{\linewidth}{llccccccccccc}
\toprule
Scenes & \multicolumn{2}{c}{Apartment 1} & \multicolumn{4}{c}{Apartment 2} & \multicolumn{4}{c}{Office 1} & \multicolumn{2}{c}{Office 2} \\
\cmidrule(r){1-1} \cmidrule(r){2-3} \cmidrule(r){4-7} \cmidrule(r){8-11} \cmidrule(r){12-13}
Method & kitchen & living & bed & kitchen & living & luke & gates362 & gates381 & lounge & manolis & 5a & 5b \\
\midrule
SCRNet~\cite{scrnet} & 2.3 / 1.3 & 2.4 / 0.8 & 3.3 / 1.5 & 2.1 / 1.0 & 4.2 / 1.8 & 4.4 / 1.4 & 2.6 / 0.8 & 3.4 / 1.4 & 2.7 / 0.9 & 1.8 / 1.0 & 3.6 / 1.5 & 3.4 / 1.2 \\
SCRNet-ID~\cite{scrnet-id} & 2.6 / 1.4 & 2.0 / 0.8 & 2.0 / 0.8 & 1.8 / 0.9 & 3.0 / 1.2 & 3.7 / 1.3 & 2.1 / 1.0 & 2.9 / 1.2 & 3.4 / 1.1 & 2.6 / 1.2 & 3.3 / 1.2 & 3.8 / 1.3 \\
NeRF-SCR~\cite{uncertainty-loc} & \fs{0.9} / \fs{0.5} & 2.1 / 0.6 & 1.6 / 0.7 & 1.2 / 0.5 & 2.0 / 0.8 & \nd 2.6 / 1.0 & 2.0 / 0.8 & \nd 2.7 / 1.2 & \nd 1.8 / \fs{0.6} & 1.6 / 0.7 & \nd 2.5 / 0.9 & \nd 2.6 / 0.8 \\
PNeRFLoc~\cite{pnerfloc} & 1.0 / 0.6 & \nd 1.5 / 0.5 & \fs{1.2} / \fs{0.5} & \fs{0.8} / \fs{0.4} & \nd 1.4 / \fs{0.5} & 8.1 / 3.3 & \nd 1.6 / 0.7 & 8.7 / 3.2 & 2.3 / 0.8 & \nd 1.1 / \fs{0.5} & X & 2.8 / 0.9 \\
Ours & \fs{0.9} / \fs{0.5} & \fs{1.1} / \fs{0.4} & \nd 1.3 / 0.6 & \nd 1.0 / 0.6 & \fs{1.2} / \fs{0.5} & \fs{1.4} / \fs{0.7} & \fs{1.1} / \fs{0.5} & \fs{1.1} / \fs{0.5} & \fs{1.7} / \fs{0.6} & \fs{1.0} / \fs{0.5} & \fs{1.3} / \fs{0.6} & \fs{1.5} / \fs{0.5} \\
\bottomrule
\end{tabularx}
\label{tab:12scenes}
\end{table*}

\section{EXPERIMENTS}

\subsection{Dataset, Baselines, and Evaluation Protocol} 
We evaluate our approach on two commonly used datasets: Replica~\cite{julian:2019:replica} and 12-Scenes~\cite{12scenes}.
The Replica dataset, which contains high-quality RGB-D sequences, is widely used in recent NeRF-based works. 
Following the setup in~\cite{uncertainty-loc}, we use 8 scenes provided by~\cite{zhi2021semantic-nerf} for evaluation. For each scene, the first sequence is used for the training set, and the second sequence is used for testing.
For the 12-Scenes dataset, we follow the common setting~\cite{12scenes}, where the first sequence is used for testing, and the remaining sequences are used for training. Instead of using all training images, we sample one frame every five frames as the training data.

We compared our approach with recent regression-based and NeRF-based approaches, including SCRNet~\cite{scrnet}, SCRNet-ID~\cite{scrnet-id}, SRC~\cite{src}, NeRF-SCR~\cite{uncertainty-loc}, and PNeRFLoc~\cite{pnerfloc}.
To measure localization accuracy, we use the commonly adopted relative rotation error and relative translation error as metrics:
\begin{equation}
  \Delta R = \arccos((\text{Tr}(R^T \hat{R}) - 1) / 2), \quad \Delta t = ||t - \hat{t}||_2,
\end{equation}
where $\hat{t}$ and $\hat{R}$ are the ground-truth translation and rotation, respectively, and $t$ and $R$ are the estimated ones.

\subsection{Implementation Details}
For each scene, we train the scene geometry, descriptor, and semantic feature branches over 10,000 iterations. The multi-resolution hash encodings consist of 16 levels of detail, with each level containing a 2-dimensional feature vector, and the finest resolution is set to 2 cm.
In the geometry branch, both the geometry and appearance decoders are 2-layer MLPs with 32 hidden units each. In the semantic branch, we use two separate 2-layer MLPs with 128 hidden units for learning descriptors and contextual features.
For SAM~\cite{sam} and SuperPoint~\cite{superpoint}, we use the `vit\_h' model and `inloc' configuration, respectively, resulting in 2D feature maps with 256 output dimensions.

\subsection{Localization Results}
In this part, we evaluate the localization performance of our proposed approach, presenting both quantitative and qualitative results.

\myvspace\noindent\textbf{Quantitative and Qualitative Results.}
We compare our approach with other baselines on Replica and 12-Scenes datasets.
Quantitative results for both datasets are shown in Tab.~\ref{tab:replica} and Tab.~\ref{tab:12scenes}, respectively.
The best localization results are highlighted in bold.
For the Replica dataset, our approach achieves state-of-the-art performance across all scenes. Since the dataset contains high-quality depth data, our SDF-based representation models the geometric information of the scene more accurately.
For the 12-Scenes dataset, our approach achieves the best localization results on 10 scenes and the second-best on the remaining two. Real-world scenes typically feature complex lighting and material conditions. While PNeRFLoc~\cite{pnerfloc} failed (denoted as ``X") on challenging scenes like ``Ofiice2/5a'', our approach performs visual localization robustly.

\begin{figure}[ht]
\centering
\includegraphics[width=0.49\linewidth]{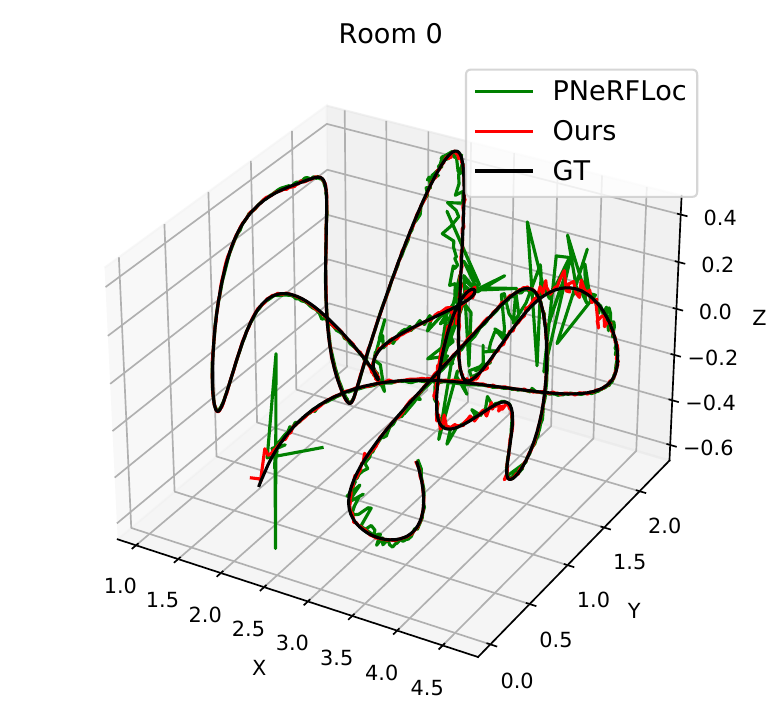}
\includegraphics[width=0.49\linewidth]{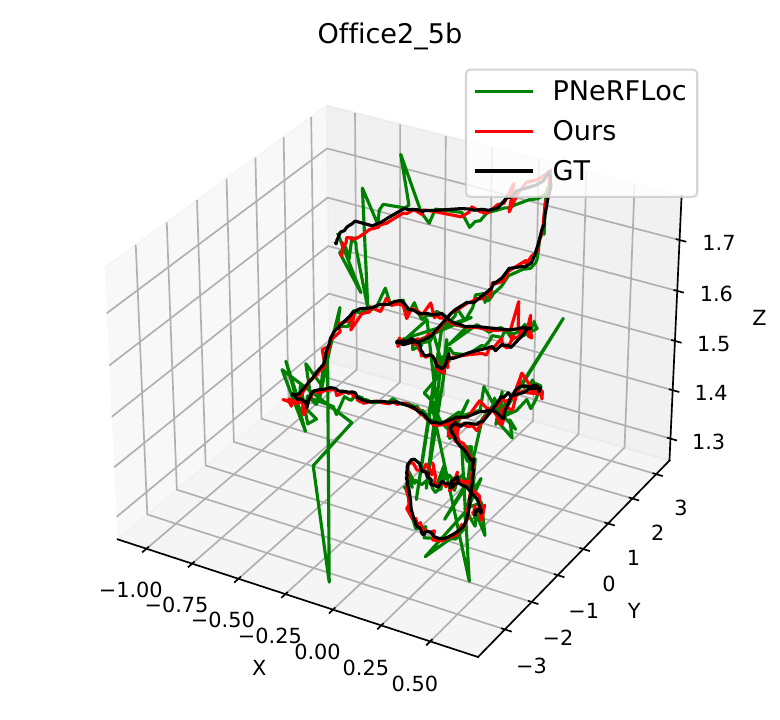}
\caption{Trajectory visualization of two selected scenes.}
\label{fig:traj}
\end{figure}


\begin{figure*}[h]
  \centering
  \scriptsize
  \setlength{\tabcolsep}{1.5pt}
  \newcommand{\sz}{0.95}
  \begin{tabular}{lc}    
    \makecell{\rotatebox{90}{12-Scenes \quad \quad \quad \quad Replica}} &
    \makecell{\includegraphics[width=\sz\linewidth]{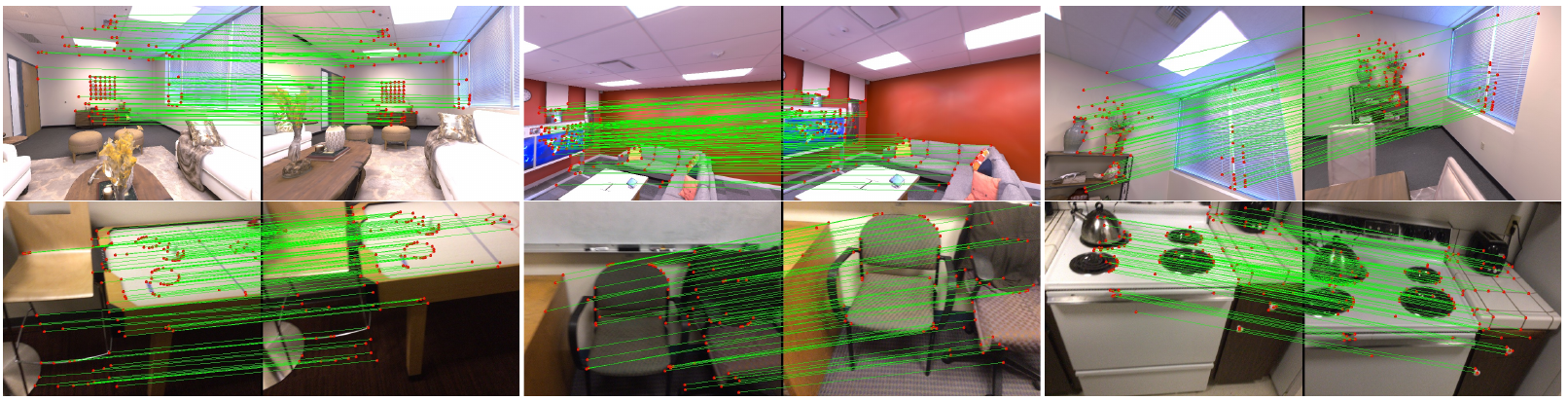}} \\
  \end{tabular}
\caption{Qualitative results of feature matching. We show some matching results of our method on Replica~\cite{julian:2019:replica} and 12-Scenes~\cite{12scenes} dataset.}
  \label{fig:matching}
\end{figure*}

We present the estimated camera trajectory of two select scenes (``Room 0'' from Replica and ``Office2/5b'' from 12-Scenes) in Fig.~\ref{fig:traj}.
Compared to PNeRFLoc~\cite{pnerfloc}, our method demonstrates more stable pose estimation with fewer outliers.
Additionally, visual matching results for both datasets are shown in Fig.~\ref{fig:matching}.
For visualization, we project the 3D map points into the reference image.
As illustrated, our approach accurately estimates correspondences despite changes in viewpoints.


\begin{table}[h]
\centering
\caption{Training time and memory usage of different methods.}
\setlength{\tabcolsep}{5pt}
\begin{tabularx}{0.93\linewidth}{lcccc}
\toprule
Method & SCRNet & NeRF-SCR & PNeRFLoc & Ours  \\
\midrule
Training Time & 2 days & 16 hours & 1 hours & 20 mins \\
Memory & 165 MB & - & 788 MB & 15.48 MB \\
\bottomrule
\end{tabularx}
\label{tab:time_memory}
\end{table}

\myvspace\noindent\textbf{Training Time and Memory Usage.}
In Tab.~\ref{tab:time_memory}, we show the training time and memory usage of different methods in scene ``Office2/manolis''.
Both PNeRFLoc and our approach were tested on an AMD Ryzen 9 7950X 16-core CPU and an RTX 4090 24GB GPU, while the results of other methods are taken from their respective papers.
As shown, regression-based approaches generally require fewer parameters but take longer to train and are less effective for localization compared to feature-based methods. Our method strikes a balance, achieving better localization performance with fewer parameters and faster convergence.

\begin{figure}[ht]
\centering
\includegraphics[width=0.49\linewidth]{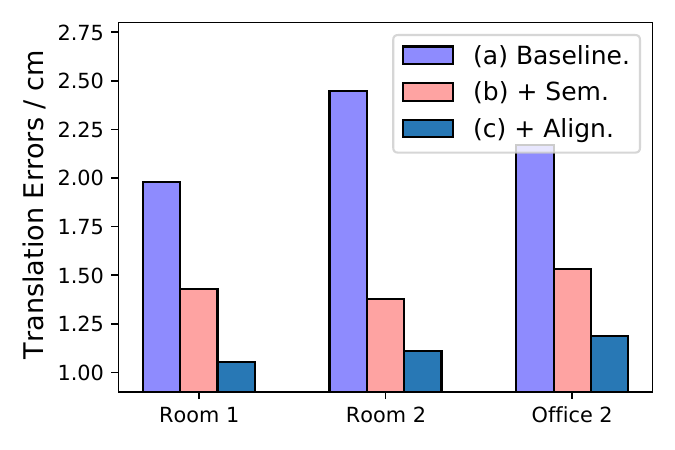}
\includegraphics[width=0.49\linewidth]{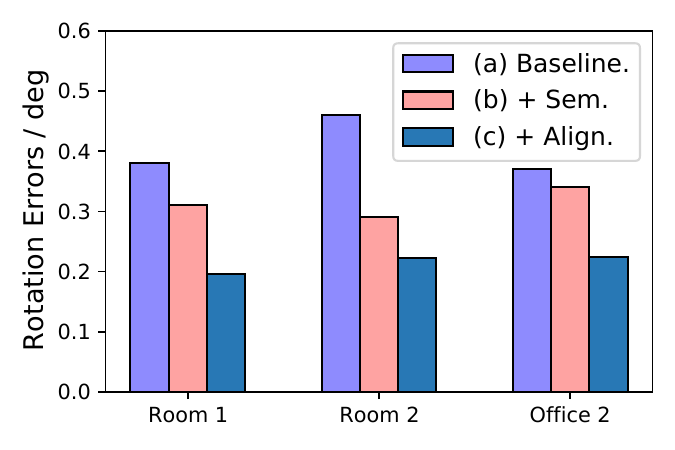}
 \caption{Median localization errors (cm, degree) of using different features.}
\label{fig:ablation_rt}
\end{figure}

\subsection{Ablation Studies}
In this part, we perform ablation studies to investigate the effect of components and designs in our system.

\myvspace\noindent\textbf{Effects of Different Features and Alignment.}
Here, we show the localization performance using different features and alignment loss for localization.
(a) Directly matching learned descriptors with nearest neighbor search.
(b) Using semantic contextual features to construct the matching graph.
(c) Using similarity alignment loss to reduce the gap between 2D and 3D feature space.
The comparison results of different settings are shown in Fig.~\ref{fig:ablation_rt}.
As shown, incorporating semantic information in the matching graph effectively filters out keypoints with semantic ambiguity during feature matching. Additionally, using similarity alignment constraints further minimizes the domain gap between the 2D and 3D descriptor feature spaces.


\begin{table}[h]
\centering
\caption{Median Localization errors (cm, degree).}
\setlength{\tabcolsep}{9pt}
\begin{tabularx}{\linewidth}{lccc}
\toprule
Case & Description & Room 0 & Office2/5b \\
\midrule
\#1 & w/o Separate Encoding &  5.34 / 2.77 & 3.6 / 2.1  \\
\#2 & w/o Match Candidate Filter &  0.78 / 0.20 & 1.8 / 0.7  \\
\#3 &  w/o Graph Match &  1.02 / 0.69 &  2.9 / 1.1  \\
\#4 &  Ours Full &  \textbf{0.51} / \textbf{0.08} & \textbf{1.5} / \textbf{0.5}  \\
\bottomrule
\end{tabularx}
\label{tab:ablation}
\end{table}

\myvspace\noindent\textbf{Effects of Design Choices.}
In Tab.~\ref{tab:ablation}, we show the localization performance of various design choices.
The results in the table have validated the effectiveness of our system's design.
\#1 shows that without additional parametric encoding for the semantic branch, it can lead to significant performance degradation.
This is due to that a single MLP can not fit the high-frequency feature very well for large indoor scenes.
\#2 shows that using semantic contextual features to filter outlier matches can lead to accurate 2D-3D correspondence estimation.
\#3 shows that using descriptor and semantic contextual features to construct the matching graph in Sec.~\ref{subsec:localization} is also effective.

\section{CONCLUSIONS}
In this paper, we presented an efficient and novel visual localization approach based on our reconstructed neural implicit map.
Specifically, to enforce geometric constraints while minimizing storage requirements, we learn a 3D descriptor field instead of storing descriptors for individual 3D points.
Additionally, we learned a complementary semantic contextual feature field for more robust matching graph reconstruction.
Besides, to reduce the domain gap between the 2D and 3D feature spaces, we employed similarity distribution alignment, which enhances the estimation of 2D-3D correspondences.
Currently, a key limitation of our approach is its inability to scale for large-scale scene localization. Despite this, our approach offers an efficient and novel solution to visual localization, paving the way for future research and improvements in this area.

\newpage
\bibliographystyle{ieeetr}

\end{document}